# Goal-Directed Design Agents: Integrating Visual Imitation with One-Step Lookahead Optimization for Generative Design

**Ayush Raina**
Department of Mechanical Engineering
Carnegie Mellon University
Pittsburgh, PA, USA
araina@andrew.cmu.edu

**Lucas Puentes**
Department of Mechanical Engineering
The Pennsylvania State University
University Park, PA, USA
lucaspuentes22@gmail.com

**Jonathan Cagan**
Department of Mechanical Engineering
Carnegie Mellon University
Pittsburgh, PA, USA
cagan@cmu.edu

**Christopher McComb**
School of Engineering Design, Technology, and Professional Programs
The Pennsylvania State University
University Park, PA, USA
mccomb@psu.edu

**ABSTRACT**

*Engineering design problems often involve large state and action spaces along with highly sparse rewards. Since an exhaustive search of those spaces is not feasible, humans utilize relevant domain knowledge to condense the search space. Previously, deep learning agents (DLAgents) were introduced to use visual imitation learning to model design domain knowledge. This note builds on DLAgents and integrates them with one-step lookahead search to develop goal-directed agents capable of enhancing learned strategies for sequentially generating designs. Goal-directed DLAgents can employ human strategies learned from data along with optimizing an objective function. The visual imitation network from DLAgents is composed of a convolutional encoder-decoder network, acting as a rough planning step that is agnostic to feedback. Meanwhile, the lookahead search identifies the fine-tuned design action guided by an objective. These design agents are trained on an unconstrained truss design problem that is modeled as a sequential, action-based configuration design problem. The agents are then evaluated on two versions of the*

*problem: the original version used for training and an unseen constrained version with an obstructed construction space. The goal-directed agents outperform the human designers used to train the network as well as the previous objective-agnostic versions of the agent in both scenarios. This illustrates a design agent framework that can efficiently use feedback to not only enhance learned design strategies but also adapt to unseen design problems.*

1.  **INTRODUCTION**

Design is a complex, multi-step process that involves reasoning, creativity, planning, and efficient search, among other skills. Humans can visualize future states and goals using experience or domain knowledge, which helps in identifying promising search directions early in the design process [1–3]. Such abilities allow humans to radically shrink the design search space [4]. Recently, approaches that combine aspects from deep learning and lookahead search-based optimization have proven to be very successful [5–7]. These approaches are complementary. Deep learning can efficiently represent complex multi-dimensional data and also learn highly non-linear relationships from it. Meanwhile, lookahead search provides fine-grained control by optimizing explicit objectives and more generalizability, since it does not depend on observational data. This combined approach holds great potential for generative design, as design problems characteristically involve large state-action spaces and often have sparse or delayed feedback. This work presents a goal-directed design agent framework that models a human-inspired exploration methodology by leveraging visual intuition from observational data and optimizing low-level action strategies using lookahead search.

These agents are designed to solve parameterized configuration design problems [8] through decomposition as a sequential decision-making process. The goal is to identify an optimal configuration of parametric components with respect to a set of constraints and objectives using sequential decisions of adding, removing, and changing the size of components. Since a design begins as an empty set and is built iteratively to completion, feedback on quality of design does not exist when the design incomplete. This challenge of delayed feedback is similar to most real-world design problems, where the designers are guided by intuition in the initial phase of a design process. To address this challenge, the proposed agent framework splits the design process into two parts based on the availability of feedback. When the feedback is unavailable, the agent simulates the intuition by visually imitating human designers using a deep learning framework defined in [9] and a mechanism of heuristic guidance [10], which are complementary and objective-agnostic. Alternatively, when the feedback is available, the agent uses lookahead search to make designs, allowing better versatilility and more focus on explicit goals.

The proposed agent framework models this goal-directed behavior of humans using a search-based approach. Making decisions that optimize for specific objectives constitutes a critical skill in problem solving [11]. This work investigates the effect of integrating data-driven strategies with explicit objective definitions on generative design performance. The agents employ a generic lookahead search that augments the learned strategies. These agents are tested on both the original design problem and an unseen problem in which an obstacle is introduced in the construction space. The performance results show the efficacy of the goal-directed agent framework and a synergistic integration of different methodologies from deep learning, optimization, and heuristics with promising applications to design problem solving.

The rest of this note is organized as follows. Section 2 introduces the basic DLAgent architecture and the concept of heuristic guidance, along with other relevant background literature. Section 3 focuses on detailing the goal-directed agent framework. Section 4 explains the experimental setup and the human design dataset, used as a baseline and for training the visual suggestion network. Section 5 illustrates the performance results and relevant discussion. Lastly, Section 6 provides conclusions to this work.

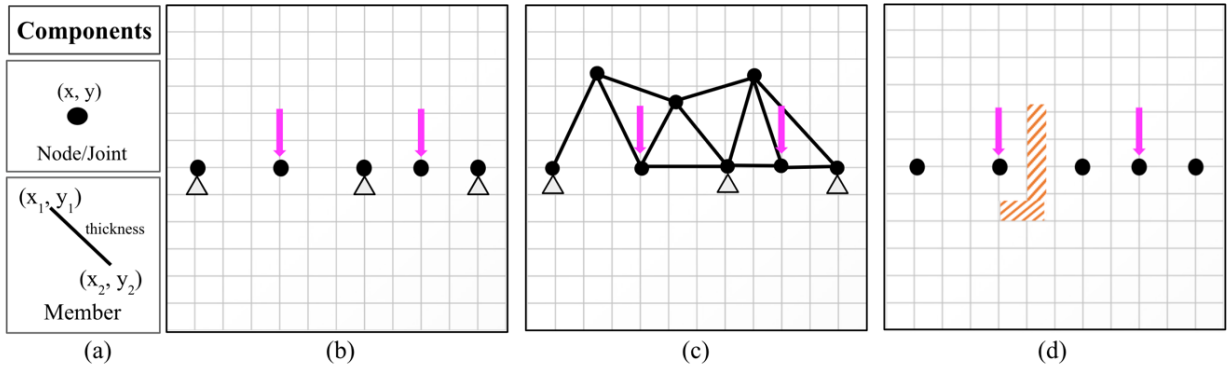

Figure 1. Truss design problem. (a) Illustrates the basic components: nodes and members that compose a truss design. (b) Initial state of the unconstrained truss design problem. (c) An example truss design. (d) Initial state of the constrained problem with orange hashed area representing the obstacle.

## 2 BACKGROUND
### 2.1 DLAgents and Heuristic Guided DLAgents

Deep Learning Agents, or DLAgents, were introduced by Raina et al. [9] as a framework to implicitly capture design strategies from an image-based dataset of design state progression. These agents achieved human-level performance in truss design without any explicit information about the objective. There are two components of the agent framework: a deep learning network and an inference algorithm. The deep learning network creates a low-dimensional representation of prior design states and uses that to envision future states towards some arbitrarily-defined objectives. This visualization of future design states is similar to how humans develop mental models [12] and visualize goal states to solve the required objectives [13,14]. Once the prediction is made, the inference algorithm identifies a particular action as suggested by structural similarity [15] comparison. These DLAgents can generate meaningful truss designs and achieve performance similar to the humans used to train the network without ever performing an evaluation or receiving feedback. These agents are referred to as Vanilla DLAgents and used as a comparative baseline in this note.

To expand the decision making capabilities of Vanilla DLAgents, Puentes et al. [10] incorporated a guidance method to the framework that allows agents to follow multi-step design heuristics, experience-derived strategies that help focus on achieving specific design goals or subgoals [16]. Heuristics can improve the efficiency of design space exploration [17–19]. These new heuristic-guided agents, referred to in this note as Temporal DLAgents, select design actions to perform in a two-stage process. First, a set of candidate actions from the Vanilla DLAgent's inference algorithm gets classified as a heuristic. This classification is based on similarity to a predefined set of heuristics. Second, the candidate list gets filtered based on this heuristic classification, and the action that best continues the heuristic is selected. The heuristic is enacted for a preset number of design actions, referred to as the burst length [20], after which a new classification is performed. This heuristic guidance enhances the performance of Vanilla DLAgents [10].

However, neither Vanilla DLAgents nor Temporal DLAgents utilize real-time objective feedback. The combination of data-driven models with real-time search systems has been successful in other domains [21–25]. The current work provides a mechanism to utilize real-time objective information on the design state using a one-step lookahead search to augment the design agents. The agent framework presented in Section 3 introduces a unique combination of data-driven strategies, pre-determined temporal heuristic relationships, and an objective-based lookahead search as a generic framework for design agents.

## 2.2  Truss design problem

The design of truss structures is a classical problem in structural mechanics and engineering. A truss can be geometrically defined as a spatial arrangement of nodes and members, as shown in Figure 1(a). In this work, the problem is represented as a parameterized configuration design problem. It is further decomposed as a sequential decision-making process, as defined in previous works [9,26] in which designers sequentially select actions such as *add a node or member*, *delete a node or member*, and *increase or decrease thickness of members*. Figure 1(b) shows the initial state of the design problem used in this work, with the nodes marked with (pink) arrows representing the loading points, while the other nodes with (grey) triangles represent the supporting points. Figures 1(b) and (d) show the initial boundary conditions of the unconstrained and constrained design scenario, respectively.

## 3  Framework for goal directed design agents

This section details the framework of the proposed goal-directed design agent (see Figure 2). The framework is organized into three main parts. The first two parts represent the deep learning network and inference algorithm, which are identical to Vanilla DLAgent framework [9]. The deep learning network is trained to predict the future state of the design visually, incrementally evolving the current design state towards an implicit goal.

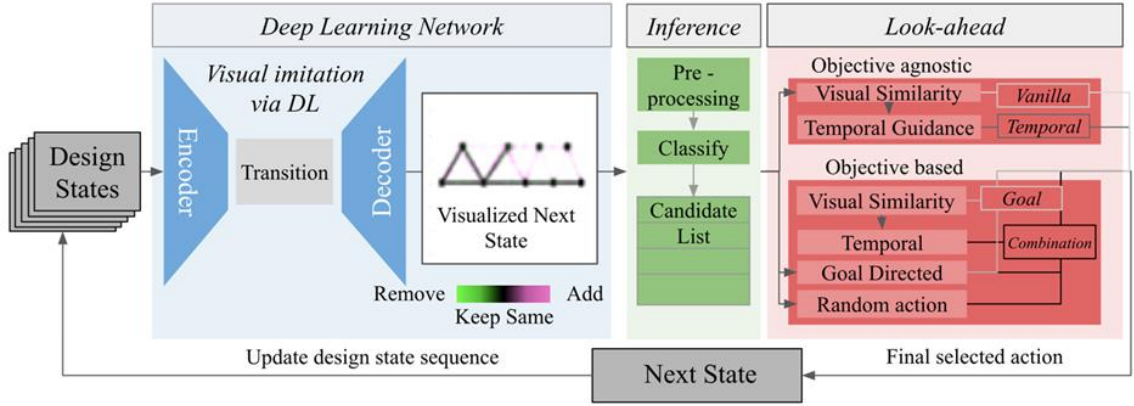

Figure 2. Overall design agent framework

The future state, represented as a heatmap, contains green and pink regions that correspond to a prediction of removing or adding material in those regions respectively. Generation of the heatmap is followed by the inference algorithm that uses rules based on image processing to identify a set of suggested candidate actions. This process maps the intuition suggested by the heatmap to a feasible set of actions in the given state. More details of these phases are provided in [9]. The final part of the framework uses lookahead evaluation, in which the agent makes the final action decision based on certain criteria and executes the selected actions to transition into a new state. This phase differentiates the new goal directed agents from the prior Vanilla DLAgents and Temporal DLAgents, as they use different selection methods.

Vanilla DLAgents use a visual similarity-based metric that maintains high similarity between the heatmap and the next state, therefore mimicking human strategies from the dataset. Alternatively, Temporal DLAgents use the additional integrated heuristics to filter candidate lists and maintain multi-step action relations. Both methods are objective function-agnostic and have no explicit information about the goal of the design process. Rather, they are dependent on the visual progression of the design states, and therefore are specific to the original problem they were trained on. In contrast, the new goal-directed agents make their action selection based on feedback from an objective function. However, like most design problems for truss design, the objective value is not defined for incomplete designs. As a result, only when a feasible design (feasibility is defined in section 4) is established can the objective-based decision making can be used. If the design is not feasible, the agent follows objective-agnostic methods. All variants of DLAgents are shown in a tabular format in Table 1 to highlight their differences.

Table 1. DLAgent variants

|  | **Objective agnostic** | **Objective guided** |
|---|---|---|
| **No heuristic guidance** | Vanilla DLAgents | Goal DLAgents |
| **Heuristic guidance** | Temporal DLAgents | Combination DLAgents |

The two new versions of the goal-directed agents are defined and compared in this note:

1. Goal DLAgents: These agents use the Vanilla DLAgent framework from the beginning of the process until a feasible state is reached. Once the objective feedback is available, a one-step lookahead search allows the agent to select the candidate action that leads to the state with the maximum objective value. This is a generic greedy method based on an arbitrary objective function and can be extended to an *N*-step lookahead evaluation. Extending to a higher number of steps should theoretically continually increase the performance at the cost of computation power, independent of the specific problem. In this work, the lookahead depth is limited to only one step. This simulates a greedy search based on a candidate list generated using a visual imitation-based prediction network. These agents isolate the effect of introducing goal directedness when compared with the baseline DLAgents.

2. Combination DLAgents: These agents use the Temporal DLAgents framework for the infeasible design phase. Once a feasible design is achieved, these agents use a combination of three methods to make the selection. The first option selects a candidate action based on heuristic guidance, allowing it to behave similarly to a Temporal DLAgent. The second option selects the action with the highest objective value, as done by Goal DLAgents utilizing the available real-time feedback. The final option selects a random candidate action to introduce some additional stochasticity to the process. This tripartite selection procedure resembles an $\epsilon$-greedy policy [27], where agents select random action with a small probability of $\epsilon$ and select the greedy strategy with a probability of $1 - \epsilon$. Here the agent selects one of these three methods based on a predefined weighting parameter tuned to balance exploration and exploitation. These combination agents illustrate that combinations of different methodologies can be integrated together to develop variations within the goal-directed framework.

## 4 EXPERIMENTAL SETUP AND DATASET

The data used for training the DLAgents is derived from a human subjects truss design study in which teams of university engineering students completed a truss configuration problem [26]. The task was to create a truss design with minimal mass while meeting a factor of safety (FOS) requirement, with a value greater than 1.0 correlating to a feasible design that will not collapse under its load. The data from the human subjects was collected through a computer-based design interface where every action was recorded. Figure 1(c) shows an example of a truss design from the data. The dataset includes human trajectories from two different problems: an unconstrained construction space, as shown in Figure 1 (b), and a constrained construction space in which an obstacle was introduced, as shown in Figure 1(d). Previously, this data has been successfully used to extract and represent design strategies as probabilistic models [28–30].

In order to maintain a fair comparison with the human designers, certain modifications are made to the agent setup. The original design study was conducted in teams of three human designers. The team members could share their experimental designs through the interface and collectively reach a final design. The experiment included sixteen teams, with each subject averaging 250 and 170 actions (or design iterations) for the unconstrained and constrained design scenarios, respectively. It should be noted that subjects average a lower number of actions in the constrained scenario, as they start from their previous design states. Further, they interacted on average after every 48 iterations. In order to simulate a naive multi-agent setup, for every team, three identical instances of the design agent are initialized and collaborate with similar interaction frequencies over a similar run of iterations.

The two new versions of the agents are tested and compared to the previously-examined agents. The prior objective-agnostic agents are Vanilla DLAgents and Temporal DLAgents. The new objective-driven agents are Goal DLAgents and Combination DLAgents. These agents are evaluated on both the constrained and unconstrained versions of the problem. All agents begin with a random configuration of truss nodes and then iteratively act to reach their final design. The agents work on their design independently unless they interact, in which case all agents always select and continue to iterate on the current highest quality design from the team's design pool. The whole process repeats until the agents reach 250 iterations for the unconstrained scenario or 700 iterations for the constrained scenario. A higher maximum iteration count is allowed for the constrained scenario since, unlike humans, agents start from scratch. However, agents are allowed to finish early in the case that no actions are suggested. As the data-driven part of these agents is only trained on the unconstrained version, the constrained version acts as an unseen problem. Their performance is illustrative of the robustness of the learned design strategies and lookahead search. In order to incorporate this constraint in the agent decision making, the candidate list generation is updated to filter out actions that violate the obstacle constraint. Implementing this filtering in the algorithm allows the design agents to respect this additional constraint without re-training the visual imitation algorithm, demonstrating the potential adaptability of the approach.

## 5  RESULTS AND DISCUSSION
### 5.1  Unconstrained construction space

Figure 3(a) shows the progression of the design agents in terms of Refined Strength-to-Weight Ratio (RSWR), which shows the SWR of only feasible designs (FOS ≥ 1.0). Specifically, this plot shows the mean of the best RSWR values reached by the agents up to a particular iteration, representing the average design progression trends. The Goal DLAgents perform better than humans and our previous baselines of Vanilla and Temporal DLAgents, but the Combination DLAgents ultimately perform the highest. This demonstrates that goal directedness greatly increases the design performance when compared with objective-agnostic methods. Further integrating heuristic guidance enhances this performance even more. The greedy solution search of Goal Agents possibly limits their ability to search for diverse designs in the full solution space. By allowing agents to decide on actions using all three selection methods, it becomes possible for them to explore widely as well as optimize greedily. Figure 3(b) and 3(c) shows the mean of the maximum value achieved for FOS and RSWR. These two metrics help in understanding the quality of the design produced. The FOS bars indicate that objective-agnostic agents (Vanilla and Temporal DLAgents) reach very high strength values, showing that they continue to add more mass to the designs exceeding FOS values of 1.0. Alternatively, Goal DLAgents have very small deviations, suggesting that once they reach the feasible threshold, they shift their focus to reducing mass and optimizing the design while maintaining a FOS around 1.0. The RSWR plots show similar trends as Figure 3(a), with Combination DLAgents performing the best.

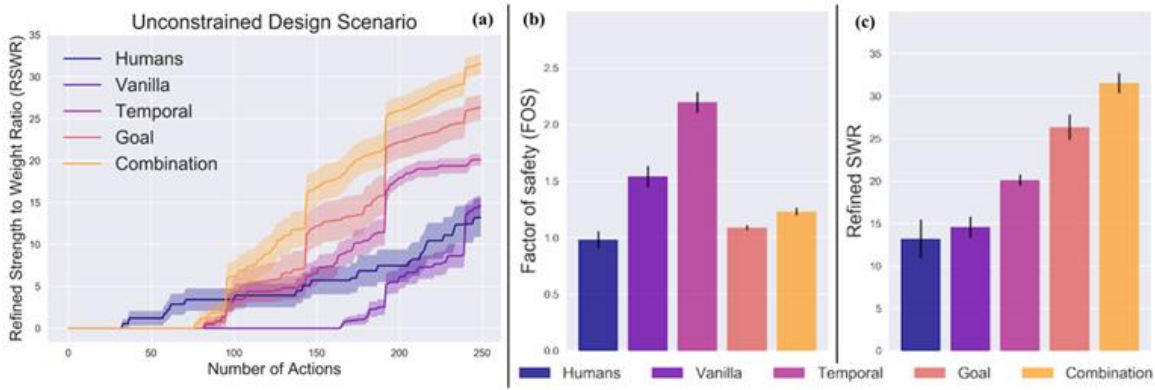

Figure 3. Unconstrained problem: (a) RSWR comparison, Bar plots for (b) FOS and (c) RSWR. ± 1 standard error indicated.

### 5.2 Constrained construction space

Figure 4(a) shows the RSWR progression of design states for the constrained problem with an obstacle (shown in Figure 1(d)). These results demonstrate that humans achieve feasible and high-performing designs early on in the process, while the agents take longer to build feasible designs. This behavior is in part due to the fact that humans begin with existing designs states (a result of the sequential nature of the original study) and modify them to adapt to the obstacle. Meanwhile, the agents begin from scratch and determine feasible solutions after approximately 200 iterations. Although both objective-agnostic agents perform similar to humans, objective-driven agents (Goal and Combined DLAgents) perform significantly better than them, illustrating the performance boost of utilizing the goal directedness on a new problem. The agents are able to outperform humans by building from their truss design strategies and further augmenting with a lookahead search. Further, it can be observed from Figure 4(a) and 4(b) that employing heuristic guidance (Combination and Temporal DLAgents) allows the agents to reach higher performance faster. In terms of FOS, it can be observed in Figure 4(c) that objective-agnostic agents continue to improve strength, often having FOS greater than 1.0, rather than focusing on efficiency by reducing mass. Finally, the RSWR bar graph in Figure 4(d) shows that humans, Vanilla, and DLAgents reach similar levels, therefore demonstrating the effectiveness of the visual imitation strategies across problems. However, these agent are bested by Goal and Combination DLAgents. Showing that Goal DLAgents exhibit opportunistic behavior, they are able to select the best performing action from their candidate list that eventually leads to high-performing designs. This indicates an effective methodology for utilizing real-time feedback information to improve design performance significantly and adapt to unseen problem constraints.

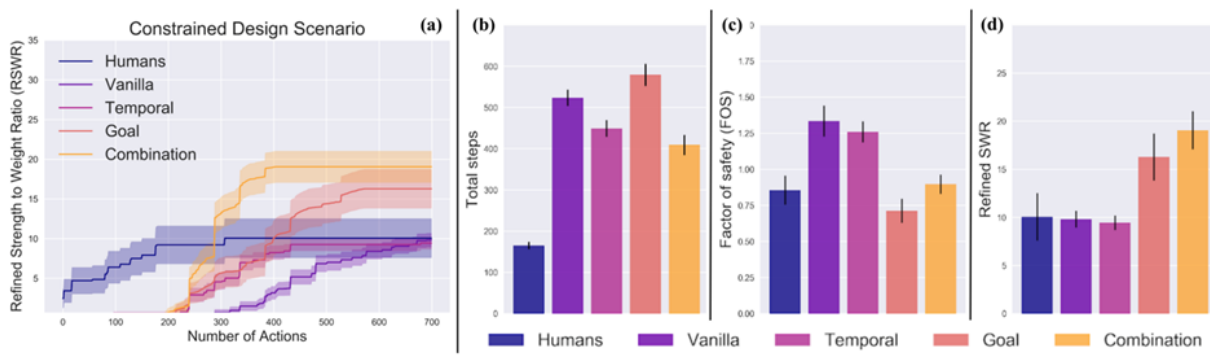

Figure 4. Constrained problem: (a) RSWR comparison (b) total steps taken, Bar plots for (c) FOS and (d) RSWR. ± 1 standard error indicated.

## 6 CONCLUSION

Humans employ a combination of high-level visual intuition and low-level control strategies that help them to efficiently navigate the search space when engaged in design tasks. Our proposed agent framework models this behavior through an autoencoder-based visual imitation learning process that guides design actions and is further optimized by a one-step lookahead search. In this work, new variations of DLAgents are explored that follow design objective-driven actions. These goal-directed agent variations demonstrate on average better performance than humans on two versions of a truss design problem. These results show that the agent can learn and then enhance design strategies from observational data. The experimental setup includes a new constrained problem on which the agents had not been trained. Currently, the agents use a rule-based inference algorithm to map the pixel-based visual guidance to actions. Future work may focus on learning to extract the feasible set of actions, and therefore developing an end-to-end trainable design agent framework.

**ACKNOWLEDGMENTS AND DISCLOSURE OF FUNDING**

This material is based upon work supported by the Defense Advanced Research Projects Agency through cooperative agreement No. N66001-17-1-4064. Any opinions, findings, and conclusions or recommendations expressed in this paper are those of the authors and do not necessarily reflect the views of the sponsors.

**REFERENCES**

[1]  Ferguson, E. S., 1994, *Engineering and the Mind's Eye*, MIT Press, Cambridge, Mass.

[2]  Kosslyn, S. M., Pascual-Leone, A., Felician, O., Camposano, S., Keenan, J. P., Thompson, W. L., Ganis, G., Sukel, K. E., and Alpert, N. M., 1999, "The Role of Area 17 in Visual Imagery: Convergent Evidence from PET and RTMS," Science (80-. )., **284**(5411), pp. 167–170.

[3]  Kosslyn, S. M., and Shwartz, S. P., 1977, "A Simulation of Visual Imagery," Cogn. Sci., **1**(3), pp. 265–295.

[4]  Brown, D. C., and Chandrasekaran, B., 1986, *Design Problem Solving: Knowledge Structures and Control Strategies*, Elsevier Science.

[5]  Silver, D., Huang, A., Maddison, C. J., Guez, A., Sifre, L., van den Driessche, G., Schrittwieser, J., Antonoglou, I., Panneershelvam, V., Lanctot, M., Dieleman, S., Grewe, D., Nham, J., Kalchbrenner, N., Sutskever, I., Lillicrap, T., Leach, M., Kavukcuoglu, K., Graepel, T., and Hassabis, D., 2016, "Mastering the Game of Go with Deep Neural Networks and Tree Search," Nature, **529**, p. 484.

[6]  Anthony, T., Tian, Z., and Barber, D., 2017, "Thinking Fast and Slow with Deep Learning and Tree Search," Adv. Neural Inf. Process. Syst., **2017**-**Decem**(Il), pp. 5361–5371.

[7]  Lee, K., Kim, S.-A., Choi, J., and Lee, S.-W., 2018, "Deep Reinforcement Learning in Continuous Action Spaces: A Case Study in the Game of Simulated Curling," J. Dy, and A. Krause, eds., PMLR, Stockholmsmässan, Stockholm Sweden, pp. 2937–2946.

[8]  Wielinga, B., and Schreiber, G., 1997, "Configuration-Design Problem Solving," IEEE Expert. Syst. their Appl., **12**(2), pp. 49–56.

[9]  Raina, A., McComb, C., and Cagan, J., 2019, "Learning to Design From Humans: Imitating Human Designers Through Deep Learning," J. Mech. Des., **141**(11), pp. 1–11.

[10] Puentes, L., Raina, A., Cagan, J., and McComb, C., 2020, "MODELING A STRATEGIC HUMAN ENGINEERING DESIGN PROCESS: HUMAN-INSPIRED HEURISTIC GUIDANCE THROUGH LEARNED VISUAL DESIGN AGENTS," Proc. Des. Soc. Des. Conf., **1**, pp. 355–364.

[11] Verschure, P. F. M. J., Pennartz, C. M. A., and Pezzulo, G., 2014, "The Why, What, Where, When and How of Goal-Directed Choice: Neuronal and Computational Principles," Philos. Trans. R. Soc. Lond. B. Biol. Sci., **369**(1655), p. 20130483.

[12] Thomas, N. J. T., 2010, "Mental Imagery > Mental Rotation (Stanford Encyclopedia of Philosophy)," pp. 1–45 [Online]. Available: http://plato.stanford.edu/entries/mental-imagery/mental-rotation.html. [Accessed: 05-Nov-2019].

[13] Athavankar, U. A., 1997, "Mental Imagery as a Design Tool," Cybern. Syst., **28**(1), pp. 25–42.

[14] Goldschmidt, G., 1992, "Serial Sketching: Visual Problem Solving in Designing," Cybern. Syst., **23**(2), pp. 191–219.

[15] Wang, Z., Bovik, A. C., Sheikh, H. R., and Simoncelli, E. P., 2004, "Image Quality Assessment: From Error Visibility to Structural Similarity," Trans. Img. Proc., **13**(4), pp. 600–612.

[16] Nisbett, R. E., and Ross, L., 1980, *Human Inference: Strategies and Shortcomings of Social Judgment*, Prentice-Hall, Englewood Cliffs.

[17] Lenat, D. B., 1983, "EURISKO: A Program That Learns New Heuristics and Domain Concepts," Artif. Intell., **21**, pp. 61–98.

[18] Laird, J., Newell, A., and Rosenbloom, P. S., 1987, "SOAR : An Architecture for General Intelligence," Artif. Intell., **33**, pp. 1–64.

[19] Langley, P., McKusick, K. B., Allen, J. A., Iba, W. F., and Thompson, K., 1991, "A Design for the ICARUS Architecture," ACM SIGART Bull., **2**(4), pp. 104–109.

[20] Königseder, C., and Shea, K., 2016, "Visualizing Relations between Grammar Rules, Objectives, and Search Space Exploration in Grammar-Based Computational Design Synthesis," J. Mech. Des. Trans. ASME, **138**(10), pp. 1–11.

[21] Schrittwieser, J., Antonoglou, I., Hubert, T., Simonyan, K., Sifre, L., Schmitt, S., Guez, A., Lockhart, E., Hassabis, D., Graepel, T., Lillicrap, T., and Silver, D., 2019, "Mastering Atari, Go, Chess and Shogi by Planning with a Learned Model," ArXiv, **abs/1911.0**, pp. 1–21.

[22] Vinyals, O., Babuschkin, I., Czarnecki, W. M., Mathieu, M., Dudzik, A., Chung, J., Choi, D. H., Powell, R., Ewalds, T., Georgiev, P., Oh, J., Horgan, D., Kroiss, M., Danihelka, I., Huang, A., Sifre, L., Cai, T., Agapiou, J. P., Jaderberg, M., Vezhnevets, A. S., Leblond, R., Pohlen, T., Dalibard, V., Budden, D., Sulsky, Y., Molloy, J., Paine, T. L., Gulcehre, C., Wang, Z., Pfaff, T., Wu, Y., Ring, R., Yogatama, D., Wünsch, D., McKinney, K., Smith, O., Schaul, T., Lillicrap, T., Kavukcuoglu, K., Hassabis, D., Apps, C., and Silver, D., 2019, "Grandmaster Level in StarCraft II Using Multi-Agent Reinforcement Learning," Nature, **575**(7782), pp. 350–354.

[23] Brown, N., and Sandholm, T., 2018, "Superhuman AI for Heads-up No-Limit Poker: Libratus Beats Top Professionals," Science (80-. )., **359**(6374), pp. 418–424.

[24] Nair, A., Pong, V., Dalal, M., Bahl, S., Lin, S., and Levine, S., 2018, "Visual Reinforcement Learning with Imagined Goals," Adv. Neural Inf. Process. Syst., **2018**-**Decem**, pp. 9191–9200.

[25] Oh, J., Guo, X., Lee, H., Lewis, R., and Singh, S., 2015, "Action-Conditional Video Prediction Using Deep Networks in Atari Games," Adv. Neural Inf. Process. Syst., **2015**-**Janua**, pp. 2863–2871.

[26] McComb, C., Cagan, J., and Kotovsky, K., 2015, "Rolling with the Punches: An Examination of Team Performance in a Design Task Subject to Drastic Changes," Des. Stud., **36**(C).

[27] Sutton, R. S., and Barto, A. G., 2018, *Reinforcement Learning: An Introduction*, A Bradford Book, Cambridge, MA, USA.

[28] McComb, C., Cagan, J., and Kotovsky, K., 2017, "Utilizing Markov Chains to Understand Operation Sequencing in Design Tasks," *Design Computing and Cognition '16*, pp. 401–418.

[29] McComb, C., Cagan, J., and Kotovsky, K., 2017, "Mining Process Heuristics From Designer Action Data via Hidden Markov Models," J. Mech. Des., **139**(11), p. 111412.

[30] Raina, A., Cagan, J., and McComb, C., 2019, "Transferring Design Strategies From Human to Computer and Across Design Problems," J. Mech. Des., **141**(11).